\documentclass[letterpaper]{article} 
\usepackage{aaai24}  
\usepackage{times}  
\usepackage{helvet}  
\usepackage{courier}  
\usepackage[hyphens]{url}  
\usepackage{graphicx} 
\usepackage{natbib}  
\usepackage{caption} 
\usepackage{algorithm}
\usepackage{algorithmic}
\usepackage{amsmath}
\usepackage{amssymb}
\usepackage{newfloat}
\usepackage{listings}
\DeclareCaptionStyle{ruled}{labelfont=normalfont,labelsep=colon,strut=off} 
\floatstyle{ruled}
\newfloat{listing}{tb}{lst}{}
\floatname{listing}{Listing}
\newcommand{\methodname}{{\tt{MIEM}}}
\title{Transformer-Empowered Multi-Modal Item Embedding \\for Enhanced Image Search in E-commerce}

\author{
    Chang Liu\textsuperscript{\rm 1,2},
    Peng Hou\textsuperscript{\rm 1},
    Anxiang Zeng\textsuperscript{\rm 2},
    Han Yu\textsuperscript{\rm 2}
}
\affiliations{
    \textsuperscript{\rm 1}Shopee Pte. Ltd., Singapore\\

    \textsuperscript{\rm 2}School of Computer Science and Engineering, Nanyang Technological University, Singapore\\
    \{chang.liulc, peng.hou\}@shopee.com, \{zeng0118, han.yu\}@ntu.edu.sg
}

\begin{document}

\maketitle

\begin{abstract}
Over the past decade, significant advances have been made in the field of image search for e-commerce applications. Traditional image-to-image retrieval models, which focus solely on image details such as texture, tend to overlook useful semantic information contained within the images. As a result, the retrieved products might possess similar image details, but fail to fulfil the user's search goals. Moreover, the use of image-to-image retrieval models for products containing multiple images results in significant online product feature storage overhead and complex mapping implementations. In this paper, we report the design and deployment of the proposed \underline{M}ulti-modal \underline{I}tem \underline{E}mbedding \underline{M}odel (\methodname{}) to address these limitations. It is capable of utilizing both textual information and multiple images about a product to construct meaningful product features. By leveraging semantic information from images, \methodname{} effectively supplements the image search process, improving the overall accuracy of retrieval results. \methodname{} has become an integral part of the Shopee image search platform. Since its deployment in March 2023, it has achieved a remarkable 9.90\% increase in terms of clicks per user and a 4.23\% boost in terms of orders per user for the image search feature on the Shopee e-commerce platform.
\end{abstract}

\section{Introduction}
The image search feature on modern e-commerce platforms has become a convenient and efficient way for users to find desired products without the need to input keywords. When a user uploads an image, this feature automatically identifies items within the picture and returns relevant products on a given e-commerce platform. This functionality not only saves time and effort for the users, but also enhances their shopping experience.
The image search feature often relies on an image embedding modeling \cite{chicco2021siamese} to extract features from images. Based on these features, K-nearest neighbors (KNN) clustering \cite{fix1989discriminatory} is often leveraged to identify similar images within the database. 

\begin{figure}[t]
\includegraphics[width=1\linewidth,page=1]{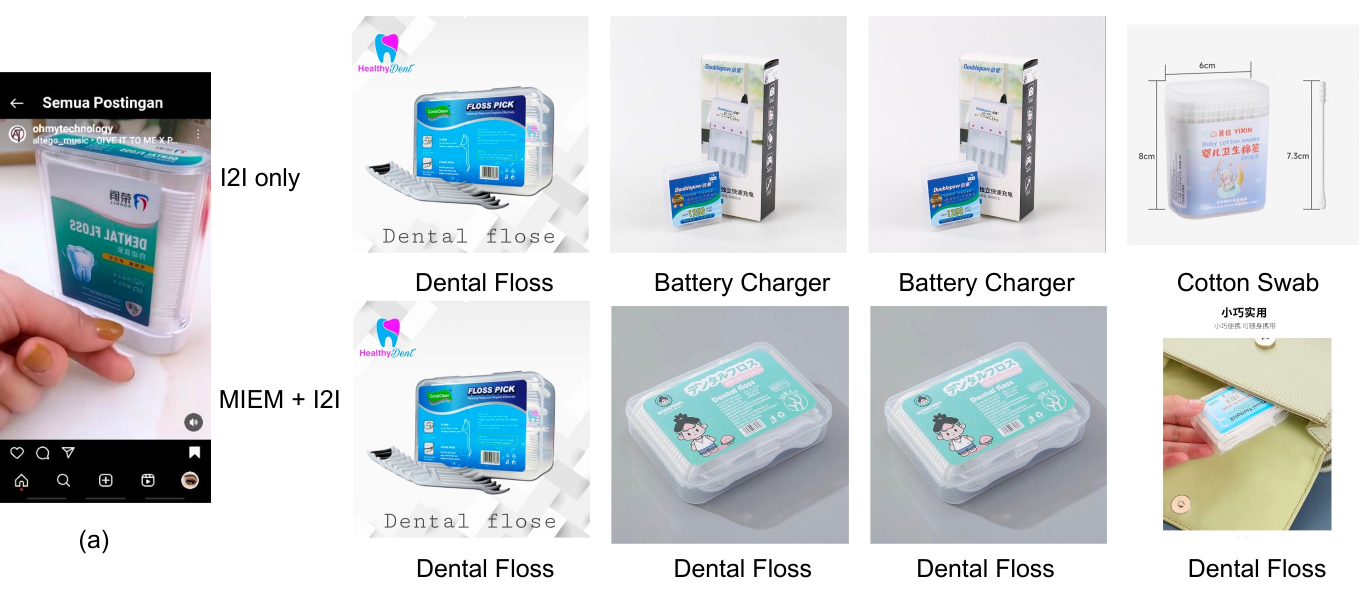}\\
\includegraphics[width=1\linewidth,page=2]{example_intro.pdf}
\caption{Search results comparison using different models. The left column displays user queries as images, while the right column showcases the top 4 retrieved items. Each item image is accompanied by its respective product name. Notably, employing Image-to-image (I2I) solely emphasizes visual details, disregarding semantic information. However, by integrating \methodname{} with I2I, these limitations can be effectively addressed.}\label{fig:example_intro}
\end{figure}

However, such an single-modal approach has two major limitations.
Firstly, it overemphasizes on visual details. The image embedding model tends to retrieve products with similar image details, but not necessarily belonging to the same category. For instance, a query for dental floss in Figure~\ref{fig:example_intro}(a) often results in products with similar packaging (e.g., battery chargers, cotton swabs) being retrieved. 
Secondly, it neglects semantic information within the image. The presence of detection model biases on the product side may introduce noises, leading to erroneous retrievals when multiple product entities exist within an image. For example, in Figure~\ref{fig:example_intro}(b), the latter two retrieved products are a plastic bottle and a necklace, with the soda can merely serving as a reference object and not a product itself. Such issues can be mitigated with multi-modal techniques, which incorporate both textual and visual information and are robust to noises.

In addition, image embedding poses engineering challenges. Due to the large number of products in Shopee's e-commerce platform\footnote{\url{https://shopee.sg/}} and the need for multiple-replications \cite{zhang2018visual}, generating an embedding for each image requires a huge index volume. Furthermore, dealing with multiple images per product requires mapping and de-duplication, further increasing engineering complexity.

Technologies that have the potential to mitigate these challenges have emerged. Vision-language pretrained models like CLIP \cite{radford2021learning} can align images and texts in the same feature space, thereby supporting user image-based product search. However, these solutions fail to effectively cross-reference information across product images and texts, resulting in low recall score during recall phase. \citeauthor{cheng2023mixer} employed concept-aware modality fusion to integrate information from a single image with texts, but it cannot accommodate multiple images \cite{cheng2023mixer}. Besides, it reliance on KnnSoftmax \cite{song2020large} for training also increases implementation complexity during training. A line of previous works (e.g., METER\cite{dou2022empirical} and X-VLM \cite{zeng2022multi}) has addressed image-to-text ranking a merge/cross attention transformer to generate matching scores for pairs of images and texts. However, these models have limitations in fusing multi-modal information during the recall phase.

To tackle these issues comprehensively, we propose the Multi-modal Item Embedding Model (\methodname{}). In this paper, we present the design and deployment experience of \methodname{} in the Shopee e-commerce platform. \methodname{} takes multiple images and product titles as the input, and leverages a Merge Attention Transformer Module \cite{dou2022empirical} to fuse information from both images and texts, thereby producing an embedding for a given product. The incorporation of product titles injects explicit semantic information into the embedding, effectively alleviating the issue of the image embedding model excessively focusing on image details. \methodname{} has been deployed in the Shopee e-commerce platform since March 2023 to power its image search business. Compared to the previous adopted solution, \methodname{}  has led to a remarkable 9.90\% increase in clicks per user and a 4.23\% boost in the orders per user metric.

\section{Application Description}
\begin{figure}[h]
\includegraphics[width=\linewidth]{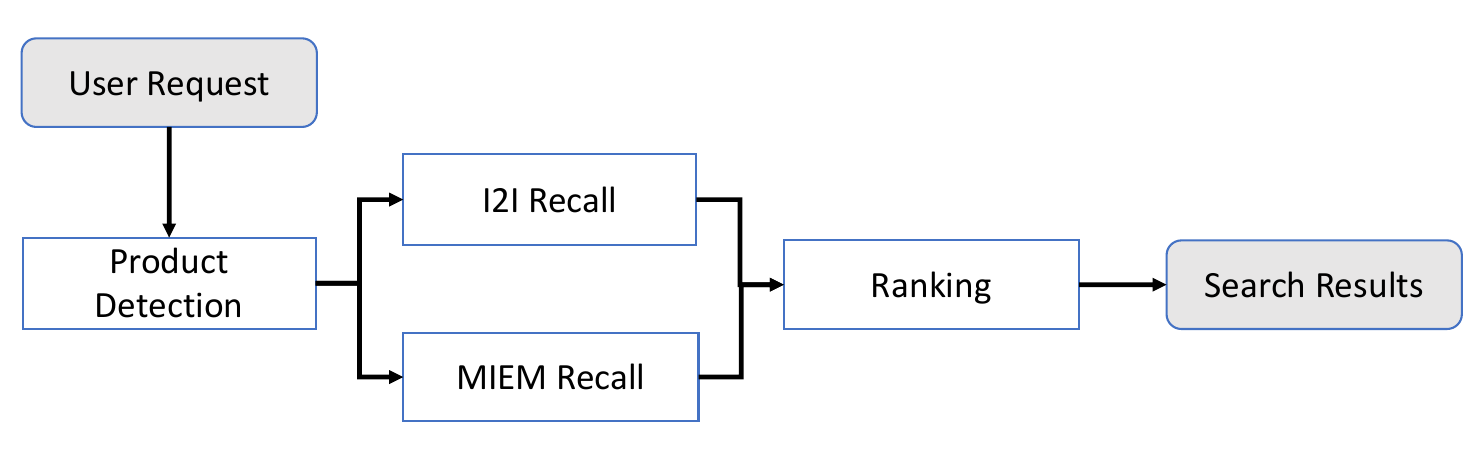}
\caption{An overview of the Shopee Image Search Engine.}\label{fig:general_framework}
\end{figure}
The overall workflow of the Shopee Image Search Engine is depicted in Figure~\ref{fig:general_framework}. The system takes a user's request (i.e., an uploaded image) as the input, and outputs a list of products it deems to be relevant. Upon receiving the user's request, the system performs the following steps: 
\begin{enumerate}
    \item It executes a product detection model to obtain the bounding boxes for the products in the image.
    \item The cropped images from the bounding boxes are then passed through two recall models: 1) Image-to-Image (I2I) recall, and 2) \methodname{} recall. 
    \item The recalled products are sent to the ranking model, where they are sorted in descending order of their ranking scores, producing the final list of recommended products.
\end{enumerate}

\begin{figure*}[h]
\centering
\includegraphics[width=1\linewidth]{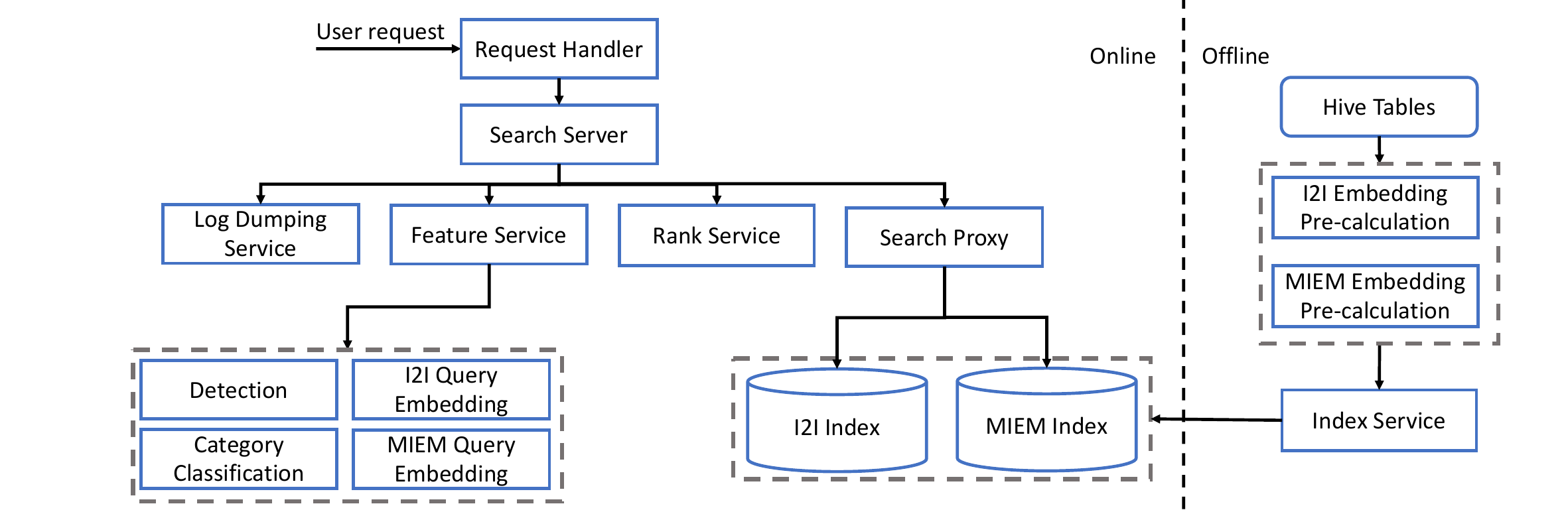}
\caption{The integration of \methodname{} into the Shopee Image Search Engine. The arrows indicate the request triggering process.}\label{fig:app_arch}
\vspace{-12pt}
\end{figure*}
The integration of \methodname{} into the Shopee Image Search Engine is based on the framework shown in Figure~\ref{fig:app_arch}, consisting of the online and the offline components. The online component performs the following steps to retrieve the products using \methodname{}:
\begin{enumerate}
    \item Upon receiving a user request, the request handler performs the initial task of parsing the request from various sources, such as webpages or different versions of the mobile app, and transforms it into a unified format. Once the request has been standardized, it is subsequently forwarded to the search server for further processing. 
    \item The search server invokes the feature service module, which then calls the detection service module to obtain the bounding boxes for the products in the image.
    \item The cropped query image is used to invoke both the I2I model and the \methodname{} model services to obtain embeddings for the products.
    \item The search proxy uses these embeddings to perform approximate nearest neighbor (ANN) search using the HNSW method \cite{malkov2018efficient} in two index databases: 1) the I2I index and 2) the \methodname{} index. For the I2I index, the corresponding product IDs are obtained for the top-$K$ retrieved images, and then subjected to de-duplication (i.e., products with same IDs will only pertain one in the recalled list). For \methodname{}, the product IDs can be directly returned. Then, the search proxy merges the two sets of retrieved product IDs and sends them back to the search server.
    \item The search server forwards the retrieved product list to the rank service, which scores each product based on relevance, popularity and other factors. The products are then sorted in descending order of their scores and returned as the final output list.
    \item The log dumping service records and stores various user-related activities, including user requests, impressions, and interactions with the search results (e.g., click, add-to-cart, order). This helps to facilitate data analysis as well as model improvements to enhance the overall user experience.
\end{enumerate}

The offline component is responsible for building and updating the product indices, and operates as a scheduled task. For the I2I embedding pre-calculation task, the system first perform product detection on the product images, and subsequently passes the cropped products to the I2I model to obtain the embeddings. For the \methodname{} embedding pre-calculation task, the system directly inputs the associated product images and titles into the model to obtain the embeddings. The embeddings from both tasks are added to their respective index database using the index service. In addition, the index service manages the removal of indices corresponding to unavailable products.

\section{Use of AI Technology}
The proposed \methodname{} approach primarily utilizes user queries and their corresponding clicked products as the training data. In this section, we introduce the model design of \methodname{} and how to train this model.

\subsection{The Design of \methodname{}}
\begin{figure*}[]
\includegraphics[width=\linewidth]{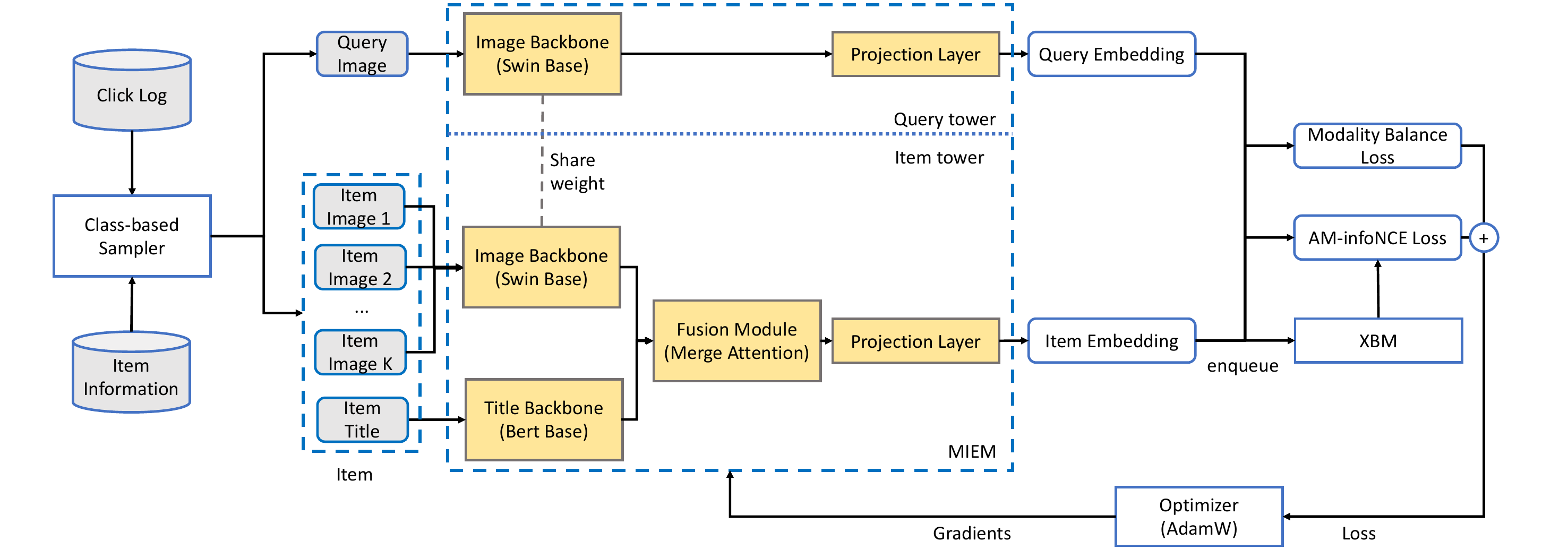}
\caption{The system architecture of the AI Engine (\methodname{}). The yellow components are the trainable parts of the entire framework. The gray components are the training data.}\label{fig:model_arch}
\end{figure*}
\methodname{}, as depicted in Figure~\ref{fig:model_arch}, adopts a dual-tower structure comprising two distinct components: the query tower and the item tower. The query tower takes the customer's uploaded query image as input and generates the corresponding query embedding as output. On the other hand, the item tower processes text and multiple product images as input and produces the item embedding. Notably, the image encoder is shared between the query and the item towers, thereby reducing the number of parameters required and improving the recall score.
The details of the model design are as follows:
\begin{enumerate}
    \item \textit{Image Encoder}: the Swin Transformer \cite{liu2021swin} is adopted to build the image encoder. It is a novel vision transformer model that employs hierarchical feature representations with linear computational complexity relative to input image size. Compared to the previous Vision Transformer (ViT) \cite{dosovitskiy2020image}, Swin Transformer has achieved state-of-the-art results in ImageNet classification, COCO object detection and semantic segmentation tasks. The Swin Transformer base is used in \methodname{} to strike a balance between accuracy and speed.
    \item \textit{Title Encoder}: following \cite{zeng2022multi}, we design the title encoder based on the multilingual BERT base \cite{devlin2018bert}.
    \item \textit{The Fusion Module}: the fusion module consists of 6 layers of merge attention networks \cite{dou2022empirical}. It concatenates patch-level embeddings from images and token-level embeddings from texts, and feeds them into a vanilla transformer model. Empirical investigations show that this structure achieves better performance than using 6 layers of cross attention (which is commonly used in image-text pretrained models). While cross attention is suitable for ranking tasks, merge attention can better preserve information from both modalities, leading to improved recall scores in modality fusion.
    \item \textit{Projection Layer}: a single fully connected (FC) layer is used as the project layer. For image encoder projection, the average of patch-level embeddings is used as the input. For fusion encoder projection, the embedding of the [CLS] token is used as the input, thereby generating the final 128-dimensional feature.
\end{enumerate}

In the training phase, we utilize search click logs to train the model. User queries are directed to the query tower, and the text and images of the clicked product are inputted into the item tower. To jointly train both towers, we employ the proposed loss functions and training methods, which will be described in detail below.

In the inference phase, the item tower computes the item embedding offline, while the query tower computes the query embedding in real-time during online queries. This setup enables image-to-product recall through vector-based approximate nearest neighbor (ANN) retrieval.

\subsection{The Training Process of \methodname{}}
A \methodname{} is trained separately for each country/region using PyTorch, and the training process involves three major steps. Following previous works \cite{zeng2022multi,li2021align,li2022blip,li2023blip}, we use AdamW \cite{loshchilov2019decoupled} as the optimizer for all three steps. 

\subsubsection{Training the Image-Text Embedding Model}
In this step of training, the fusion module is not involved. The projected query embedding and the projected text embedding are directly aligned. The widely adopted InfoNCE loss approach \cite{oord2018representation} is commonly used for this alignment:
\begin{equation}
    \mathcal{L'}(Q,T) = - \mathbb{E}_{i \in [0,N]}\left[\log\frac{e^{s(Q_i,T_i)}}{\sum_{j\in[0,N]}e^{s(Q_i,T_j)}}\right].
\end{equation}
Here, $Q_i$ denotes the query image embedding. $T_i$ represents the item title embedding. $S(Q_i, T_i)$ is the cosine similarity between $Q_i$ and $T_i$. $N$ is the batch size.

Inspired by Cosface \cite{wang2018cosface}, we propose AM-InfoNCE to further enhance the model capability. AM-InfoNCE employs two hyperparameters, $\gamma$ and $m$, to optimizing the decision boundary in the angular space by minimizing intra-class variance and maximizing inter-class variance. $\gamma$ controls the overall size of the logits, while $m$ controling the margin between classes.
\begin{equation}
\begin{split}
    \mathcal{L}(Q,T) &= - \mathbb{E}_{i \in [0,N]}[\\
                        &\log\frac{e^{\gamma (s(Q_i,T_i)-m)}}{e^{\gamma (s(Q_i,T_i)-m)}+\sum_{j\in[0,N],j\neq i}e^{\gamma s(Q_i,T_j)}}]. \label{eq:am}
\end{split}
\end{equation}

It is important to note that this loss is different from Cosface \cite{wang2018cosface}. Cosface adopts a classification loss, computing the similarity between samples in a batch and the weights of each class (product) to obtain the loss. In contrast, AM-InfoNCE uses the query as a positive sample with clicked item titles and other items in the batch as negative samples to calculate the similarity.

\subsubsection{Training with the Fusion Module}
In this step, the fusion module is involved in the training process. Only a single item image is used for training the entire model. Class-based sampling and Cross Batch Memory (XBM) \cite{wang2020cross} are used to improve performance.
Class-based sampling ensures that each minibatch contains products of the same class, providing more hard negative samples to improve model performance over random sampling.
XBM \cite{wang2020cross} is a mechanism that remembers embeddings from past iterations, thereby allowing the model to collect sufficient hard negative sample pairs across multiple batches. By connecting each anchor in the current batch with the closest embeddings in the nearest minibatch, XBM provides a large number of hard negative samples. Although XBM effectively mines hard negative samples, it is still limited by the small batch size training \cite{wang2020cross}. Thus, we avoid direct training with multiple item images.

To ensure the preservation of alignment between the image and text feature space, we propose the modality balance loss. Under this loss, the fusion embedding without the item image and texts is separately aligned with the query embedding. In this way, the model can learn more robust feature information from individual modalities. The modality balance loss is computed as follows:
\begin{equation}
\begin{split}
\mathcal{L}^{\text{balance}}(Q,I,T) &= \mathcal{L}(Q,F(I^\text{def},T))+\mathcal{L}(Q,F(I,T^\text{def})) \\
                        & +\mathcal{L}(F(I^\text{def},T),Q)+\mathcal{L}(F(I,T^\text{def}),Q)
\label{eq:balance}
\end{split}
\end{equation}
where $F$ is the fusion module. $I^\text{def}$ and $T^\text{def}$ are default item images and title embedding. Their corresponding tokens have all-zero attention values within the fusion module.

The final loss is the sum of modality balance loss (Eq.~\ref{eq:balance}), AM-InfoNCE (Eq.~\ref{eq:am}) and XBM-enhanced AM-InfoNCE loss:
\begin{equation}
\begin{split}
\mathcal{L}^\text{final}(Q,I,T) =& \mathcal{L}(Q,F(I,T))+ \mathcal{L}(F(I,T),Q) \\
    &+\mathcal{L}^\text{balance}(Q,I,T) \\
    &+\mathcal{L}(Q,F_m)+ \mathcal{L}(F(I,T),Q_m),
\end{split}
\end{equation}
where $F_m$ is the fusion embedding in the XBM and $Q_m$ is query embedding in XBM \cite{wang2020cross}.

\subsubsection{Training with Multiple Item Images}
In this step, the fusion module is finetuned with multiple item images. The loss function and training techniques used in Step 2 are maintained during this process. To facilitate batch training and inference, we set a fixed number of item images to $K$ ($K>1$). For items with an image count below $K$, we provide empty images to reach the required number of $K$. Conversely, for items with an image count exceeding $K$, we truncate the images to match the fixed number $K$.

After conducting experiments, we determined that setting $K=4$ strikes a balance between inference speed and recall accuracy. This choice allows us to efficiently process batches during training and inference while still achieving satisfactory recall performance, resulting in a practical and effective configuration for the model.

\subsection{Acquisition of Training Data}
We leverage user click logs to train \methodname{}. Each log entry is represented as a triplet $\langle$query image, clicked item title, clicked item image list$\rangle$. Here, the clicked item title and clicked item image list correspond to the titles and images of the products that users clicked after using the query image for search. While theoretically, users may click on completely unrelated products, through manual reviewing and analysis, we found that approximately $45\%$ of the triplets consist of query-item pairs that are identical, $50\%$ are on similar items, and less than $5\%$ are on unrelated noisy products. Thus, the dataset is useful for our purpose.

We conjecture that \methodname{} should be able to learn even from the similar item triplets. The reasons are as follows:
\begin{enumerate}
    \item From the perspective of the entire system, these similar items are products of interest to users. Even if they are not identical, they should be retrieved in the recall stage.
    \item From a learning perspective, although similar item triplets do not all perfectly match, often the (query image, clicked item title) pairs correspond correctly, with only differences in images. Thus, the model can still learn useful associations between the queries and the products from these examples.
\end{enumerate}
Thus, we retain this portion of the data containing similar item triplets. As for the $5\%$ of completely unrelated data, we have not conducted specific cleansing for them at present. This could be a potential point for future extension.

\section{Application Development and Deployment}
\methodname{} has been deployed in the Shopee Image Search Engine, serving customers from Southeast Asia, Taiwan and Brazil since March, 2023. The deployment follows the process described in Figure \ref{fig:app_arch}. Throughout the development and deployment process, a collection of technologies has been developed to meet practical requirements.

\subsection{Online Query Feature Extraction}
To improve the online service delivery efficiency, the image encoder (a PyTorch \cite{paszke2019pytorch} model) is converted into a TensorRT \cite{vanholder2016efficient} model and integrated into the online feature extraction service. This service takes images and bounding boxes predicted by the detection service as inputs, processes the images using the TensorRT model, and outputs the projected image embeddings. TensorRT optimizes the neural networks based on the GPU, leading to reduced service latency.

\subsection{Offline Item Feature Extraction}
Similarly, the item tower service is built on a TensorRT model. In this case, the item images no longer require bounding boxes. In addition, the service takes the item titles as inputs and outputs the projected item embeddings.

\begin{figure}[!b]
\includegraphics[width=\linewidth]{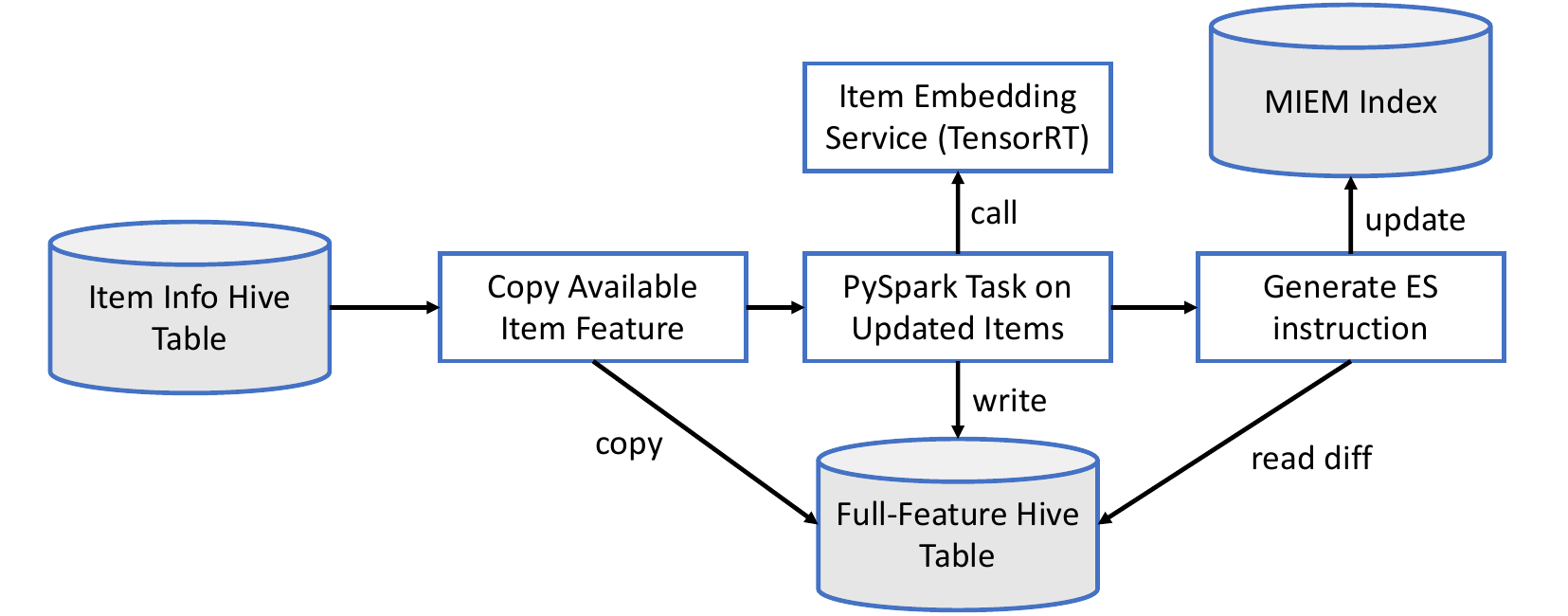}
\caption{\methodname{} item embedding pre-calculation.} \label{fig:daily_jobs}
\end{figure}

To enhance online service efficiency, we pre-calculate the embeddings for items in the database. To handle the vast size of the product database and the high number of daily addition/removal of items, we design the following pipeline (as shown in Figure~\ref{fig:daily_jobs}). We maintain a full-feature Hive table for all items, partitioned by day. Eventually, the features are inputted into the \methodname{} Elasticsearch (ES) \cite{gormley2015elasticsearch} Index. This process is executed daily as follows:
\begin{enumerate}
    \item The full-feature partition from the previous day is first copied into the current day partition. Deleted items and items with changed information (e.g., images, titles) are not copied. This is achieved by comparing the upstream product information table with the full-feature table from the previous day.
    \item The information of newly added or updated items is collected and sent to a PySpark \cite{drabas2017learning} task. This task efficiently obtains embeddings through gRPC calls \cite{wang1993grpc} to the item embedding service, and writes them into the current day partition. Items with unreadable images or titles containing only punctuation marks or spaces are not written into the partition.
    \item We generate commands to update the ES index by comparing the current day and previous day full-feature item tables. We define two actions: delete and update. The delete action removes items that exist in the previous day partition, but not in the current day partition. The update action includes newly added or updated items for which the service successfully obtained features. These commands are sent to the Index service to update the ES index \cite{gormley2015elasticsearch}.
\end{enumerate}

It is worth noting that although the full-feature Hive table appears redundant with the ES index, it is essential for two reasons. Firstly, we cannot view item features in the ES index, making debugging difficult. Secondly, due to limitations of the HNSW algorithm, the ES index requires periodic reconstruction. Otherwise, the recall score declines with the increasing number of updated items. Maintaining the full-feature Hive table facilitates the reconstruction process.

\subsection{Online Same Item Rate Analysis}
To analyze the online same item rate, we periodically use outsourced annotation to detect changes in the same item rate. The annotation task provides a query image, the top 5 product images from the search results, and the product titles. The annotators determine whether the query and the products are the same, similar or dissimilar. To facilitate annotation, we use a machine translation model to translate the product titles into the native languages of the annotators, along with the original titles. ``Same'' indicates that the query and the product are the same, while ``similar'' means that they belong to the same category (e.g., both are sweaters or dresses). Since user queries may include products that do not exist in the Shopee product catalog, evaluating the number of similar items allows us to assess whether the system includes similar products in search results when identical items are not available, thereby better satisfying users' search needs.

\section{Application Use and Payoff}

\subsection{Offline Evaluation}
\begin{table*}[!t]
\centering
\begin{tabular}{|l|r|r|r|r|r|r|}
\hline
Model            & Category Accuracy                     & Recall@1                     & Recall@5                     & Recall@10                     & Recall@50                     & Recall@100                     \\\hline
I2I              & 0.874                                 & 0.820                        & 0.922                        & 0.951                         & 0.979                         & 0.985                          \\\hline
I2T              & 0.876                                 & 0.395                        & 0.632                        & 0.714                         & 0.856                         & 0.896                          \\\hline
\methodname{}             & 0.881                                 & 0.812                        & 0.927                        & 0.952                         & 0.984                         & 0.988                          \\
\methodname{} (1 emb/img) & \textbf{0.894}                                 & 0.819                        & 0.939                        & 0.964                         & 0.988                         & 0.990                          \\
\methodname{}+I2I         & 0.879                                 & \textbf{0.872}               & \textbf{0.953}               & \textbf{0.969}                & \textbf{0.984}                & \textbf{0.985}                 \\ \hline           
\end{tabular}
\caption{The offline evaluation results on a Shopee product test set with 3 million items. The latest proposed solution of combining I2I with \methodname{} achieved the best performance.}\label{tab:offline_result}
\end{table*}

To evaluate the performance of \methodname{} at this stage, a dataset based on real-world user queries collected from the Indonesia market was constructed, containing user query images and 3 million Shopee products. Since some products may have identical items (same physical products with different IDs), a same item merging strategy was employed. For each product image $a$, the top-$K$ product images $b$ were obtained using an ANN approach. An MLP model was used to determine whether $a$ and $b$ are the same item. This MLP model takes the concatenation of the features of $a$ and $b$ as the input, and outputs the probability of them being the same item. The training data for the MLP model was annotated by external annotators.
Moreover, since the products the users click on are not necessarily the same items, these (query, clicked item) pairs were verified by external annotators to obtain accurate same item information to improve the accuracy of evaluation. A total of $9,000$ query images were retained after removing different-item queries.

Two metrics were used in the offline evaluation:
\begin{itemize}
    \item \textit{Recall@k}: commonly used in deep metric learning related papers \cite{liu2021noise,sun2020circle,wang2019multi,cakir2019deep,qian2019softtriple,ibrahimi2022learning} to determine whether the correct results are included in the top-$K$ retrieved results, Recall@k directly reflects model performance in terms of same item rate.
    \item \textit{Category Accuracy}: evaluates whether the most common category among the top 10 results matches the correct product category.
\end{itemize}

\methodname{} was compared against the following models:
\begin{itemize}
    \item \textbf{I2I}: Image-to-image retrieval with Shopee's image embedding model, based on the ViT base. To handle the issue of multiple images per product, the KNNs of the images were obtained with the duplicates removed, ensuring that only the most relevant image was kept as the representative of the product.
    \item \textbf{I2T}: A CLIP \cite{radford2021learning} model that performs image-to-text retrieval. It is the output of the first step in \methodname{} training, without the fusion encoder.
\end{itemize}

The following variants of \methodname{} were involved in offline evaluation:
\begin{itemize}
    \item \methodname{} (1 emb/img): An embedding is generated for each $\langle$image, text$\rangle$ pair about a product. This is a fair comparison with I2I.
    \item \methodname{}+I2I: Both models were used for image search, and the scores were fused as (I2I score + weight $\times$ \methodname{} score), with the weight adjustable based on performance on the test set.
\end{itemize}

The retrieval scores are reported in Table~\ref{tab:offline_result}. Compared to I2I, I2T slightly improved category accuracy with significantly reduced recall score. This is due to the ambiguity of text descriptions, making it difficult to accurately describe product textures, patterns and styles. \methodname{} achieved slightly lower Recall@1 than I2I, but outperformed it under Recall@5 and beyond. In the context of recall stage, the slight drop in Recall@1 is acceptable.
\methodname{} (1 emb/img) achieved more significant improvements in Recall@5 and beyond compared to \methodname{}, with comparable Recall@1 to I2I. It also achieved the highest category accuracy among all comparison approaches. However, since this strategy did not achieve the desired storage savings, it was not eventually selected for online deployment.
Using both I2I and \methodname{} for image search recall yielded the best results across the board, demonstrating that I2I and \methodname{} have complementary strengths. Combining both models maximized the same item rate improvement. Based on these results, the decision on deploying \methodname{}+I2I in the Shopee Image Search Engine was eventually made.

\subsection{Online Evaluation}
\begin{table*}[]
\centering
\begin{tabular}{|l|l|l|r|r|r|r|}
\hline
          & A/B Test     & Group & Clicks/User & $\Delta$Clicks/User & Orders/User & $\Delta$Orders/User \\
          & Period       &       &            &                    &            &                     \\ \hline
Shopee    & 31/03/2023   & I2I  & 3.84       &                    & 0.0804     &                     \\ \cline{3-7} 
Indonesia & --10/04/2023 & \methodname{}+I2I   & 4.22       & +9.90\%            & 0.0838     & +4.23\%             \\ \hline
Shopee    & 20/04/2023   & I2I  & 4.00       &                    & 0.0364     &                     \\ \cline{3-7} 
Brazil    & --05/05/2023 & \methodname{}+I2I   & 4.24       & +6.00\%            & 0.0382     & +4.95\%             \\ \hline
\end{tabular}
\caption{A/B test results of the Shopee Image Search Engine in Indonesia and Brazil. $\Delta$ refers to the relative improvements achieved by \methodname{}+I2I compared to I2I.}\label{tab:oneline_ab}
\end{table*}

Since its deployment in March 2023, A/B testing was conducted at Shopee's Indonesia and Brazil e-commerce marketplaces. In the online evaluation, four types of metrics were utilized:
\begin{itemize}
    \item \textit{Same \& Similar Item Recall@4}: This metric determines whether the correct results are present among the top-4 retrieved items. The determination of whether an item is considered the same or similar to the query is made by annotators, rather than relying on ground truth data used in offline evaluation. The top-4 items were assessed because, in the Shopee App, the first page of search results displays these top 4 items.
    \item \textit{Top4 Irrelevant Item Rate}: This metric evaluates the rate of irrelevant items found in the top 4 search results. Annotators are responsible for determining the relevance of items. Any items that are neither considered the same nor similar to the query are classified as irrelevant.
    \item \textit{Clicks/User}: This metric measures the number of items clicked per user within the search results. It provides insights into user engagement with the presented search results.
    \item \textit{Orders/User}: This metric measures the number of orders placed per user within the image search. It helps assess the effectiveness of the image search in driving user purchases.
\end{itemize}

The results are shown in Table~\ref{tab:oneline_ab}. It can be observed that the new I2I+\methodname{} solution for multi-modal image search recall significantly increased clicks per user and orders per user compared to the previous deployed solution of using I2I alone.
Meanwhile, according to the online similarity rate analysis method, we collected ~3,000 queries from Shopee Indonesia, and evaluated the top4 same and similarity item rates. The evaluation results in Table~\ref{tab:oneline_label} show that \methodname{} significantly increases the same and similarity item rates, whiling reducing the proportion of the top4 irrelevant products.
Furthermore, when compared to the image embedding model, \methodname{} significantly reduces the storage space requirements. In particular, for the online ES index, \methodname{} only requires a approximately $61.17\%$ less storage space compared to the previously adopted I2I index.

\begin{table}[t]
\centering
\begin{tabular}{|l|l|l|l|}
\hline
Model    & \begin{tabular}[c]{@{}l@{}}Same\\ Item\\ Recall@4 \end{tabular} & \begin{tabular}[c]{@{}l@{}}Similar\\ Item\\ Recall@4\end{tabular} & \begin{tabular}[c]{@{}l@{}}Top 4 \\ Irrelevant\\ Rate\end{tabular} \\ \hline
I2I      & 50.55\%                                                            & 93.26\%                                                               & 13.58\%                                                             \\ \hline
\methodname{}+I2I & \textbf{53.66}\%  & \textbf{95.60}\% & \textbf{10.71}\%  \\ \hline
\end{tabular}
\caption{A/B test results on Same \& Similar Item Recall@4 and Irrelevant Rate for the top 4 products from Shopee Indonesia.}\label{tab:oneline_label}
\end{table}

\begin{figure*}[t]
\includegraphics[width=\linewidth]{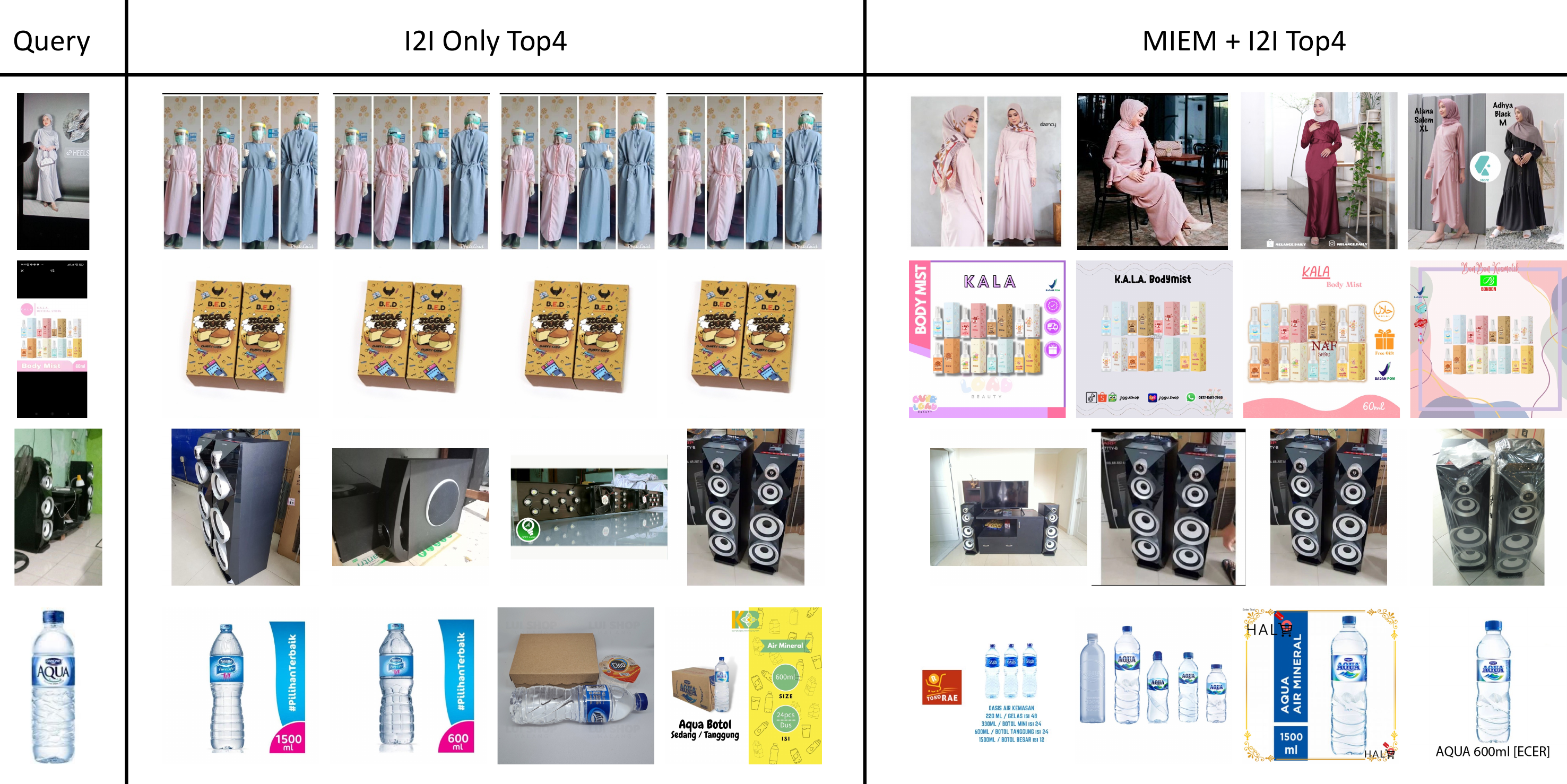}
\caption{Qualitative comparison of top4 search results using different queries under \methodname{}+I2I vs. I2I.}\label{fig:example_exp}
\end{figure*}

\subsection{Qualitative Analysis}
Figure~\ref{fig:example_exp} illustrates examples of the top 4 online cases. In the first example, the query is for Indonesian traditional attire. The I2I model mistakenly returns results for protective clothing. However, after deploying \methodname{}, this issue is significantly improved, and more relevant results are obtained. In the second example, the query is for a body spray. The I2I model returns results that resemble boxes of cakes.
\methodname{}, focusing not only on image details but also on semantic information, accurately identifies the correct product, resulting in more precise search results. 
In the third case, the query is for a speaker. \methodname{} successfully delivers top 4 results that match the appearance of the query image. In the last example, without the multi-modal approach, the I2I model retrieves an irrelevant product (the third item), mistaking the cake box for the actual product which is bottled water. The multi-modal capability of \methodname{} helps resolve such issues by considering both text and image information, leading to more accurate and contextually relevant recommendations.

\section{Lessons Learned During Development, Deployment, and Maintenance}
Throughout the development and deployment process, several lessons were learned:
\begin{enumerate}
    \item Some products are challenging for \methodname{} to handle. Products with many SKUs (e.g., phone cases with different patterns) may be difficult to cover with a single embedding. One potential solution is to create multiple embeddings for such products based on the number of SKUs or the degree of variation in the product images. Methods like softtriple \cite{qian2019softtriple} are promising solutions.
    \item In terms of deployment, achieving a balance between speed and accuracy requires the use of TensorRT. However, attention should also be paid to the pre-processing speed. In addition, the overall system outputs should be consist with the offline evaluation results.
    \item In terms of model maintenance, there are a large number of new products added to the Shopee database daily. The previously adoped I2I may be less affected by new products due to its focus on image details alone. In contrast, \methodname{}, which focuses on both the semantic information and the images, may struggle to establish relationships between texts and images for products not seen during training (e.g., newly released iPhones). Regularly retraining the model can help mitigate this issue. Nevertheless, the improvement has been deemed to outweigh the maintenance overhead incurred.
\end{enumerate}

\section{Conclusions and Future Work}
The proposed Multi-modal Item Embedding Model (\methodname{}) complements Shopee's image search by providing semantic information for both images and product texts. By considering both images and text for a product and extracting valuable information, \methodname{} constructs more context-aware feature representations for products. Compared to the previously deployed single modal image search model, \methodname{} achieves higher the recall score while requiring significantly less feature storage space compared to using Image-to-image only. The model has become an essential part of the Shopee Image Search Engine. Since its deployment in March 2023, it has help the Shopee e-commerce platform achieve $9.90\%$ higher clicks per user and a $4.23\%$ higher orders per user, thereby bringing significant benefits to its businesses.

In subsequent research, we will explore how to enhance \methodname{} from the perspective of data privacy preservation via federated learning \cite{Yang-et-al:2020FL,Goebel-et-al:2023} to further improve user experience.

\section{Acknowledgements}
This research/project is supported by the National Research Foundation Singapore and DSO National Laboratories under the AI Singapore Programme (AISG Award No: AISG2-RP-2020-019); and the RIE 2020 Advanced Manufacturing and Engineering (AME) Programmatic Fund (No. A20G8b0102), Singapore. 

\bibliography{aaai24}

\end{document}